\documentclass[11pt,a4paper]{article}
\usepackage[hyperref]{acl2020}
\usepackage{times}
\usepackage{latexsym}

\usepackage{graphicx}
\graphicspath{ {./images/} }
\usepackage{float}
\restylefloat{table}
\usepackage{amssymb}

\usepackage{microtype}
\usepackage{booktabs}
\usepackage{multirow,multicol}
\usepackage{graphicx}
\usepackage{enumitem}

\aclfinalcopy 

\title{Distilling Large Language Models into \\ Tiny and Effective Students using pQRNN}

\author{Prabhu Kaliamoorthi, Aditya Siddhant, Edward Li, Melvin Johnson \\
Google \\
\{prabhumk, adisid, gavinbelson, melvinp\}@google.com}

\date{}

\begin{document}
\maketitle
\begin{abstract}
Large pre-trained multilingual models like mBERT, XLM-R achieve state of the art results on language understanding tasks. However, they are not well suited for latency critical applications on both servers and edge devices. It's important to reduce the memory and compute resources required by these models. To this end, we propose pQRNN, a projection-based embedding-free neural encoder that is tiny and effective for natural language processing tasks. Without pre-training, pQRNNs significantly outperform LSTM models with pre-trained embeddings despite being 140x smaller. With the same number of parameters, they outperform transformer baselines thereby showcasing their parameter efficiency. Additionally, we show that pQRNNs are effective student architectures for distilling large pre-trained language models. We perform careful ablations which study the effect of pQRNN parameters, data augmentation, and distillation settings. On MTOP, a challenging multilingual semantic parsing dataset, pQRNN students achieve 95.9\% of the performance of an mBERT teacher while being 350x smaller. On mATIS, a popular parsing task, pQRNN students on average are able to get to 97.1\% of the teacher while again being 350x smaller. Our strong results suggest that our approach is great for latency-sensitive applications while being able to leverage large mBERT-like models.

\end{abstract}

\section{Introduction}
Large pre-trained language models \cite{DBLP:journals/corr/abs-1810-04805, lan2020albert, DBLP:journals/corr/abs-1907-11692, DBLP:journals/corr/abs-1906-08237, raffel2019exploring} have demonstrated state-of-the-art results in many natural language processing (NLP) tasks (e.g. the GLUE benchmark~\cite{wang-etal-2018-glue}). Multilingual variants of these models covering 100+ languages~\cite{conneau-etal-2020-unsupervised, arivazhagan2019massively, fang2020filter, siddhant2020evaluating, xue2020mt5, chung2020rethinking} have shown tremendous cross-lingual transfer learning capability on the challenging XTREME benchmark~\cite{hu2020xtreme}. However, these models require millions of parameters and several GigaFLOPS making them take up significant compute resources for applications on servers and impractical for those on the edge such as mobile platforms.

Reducing the memory and compute requirements of these models while maintaining accuracy has been an active field of research. The most commonly used techniques are quantization~\cite{gong2014compressing, han2015deep}, weight pruning~\cite{10.5555/2969239.2969366}, and knowledge distillation~\cite{hinton2015distilling}. In this work, we will focus on knowledge distillation (KD), which aims to transfer knowledge from a teacher model to a student model, as an approach to model compression. KD has been widely studied in the context of pre-trained language models~\cite{DBLP:journals/corr/abs-1903-12136, turc2019wellread, sun-etal-2020-mobilebert}. These methods can be broadly classified into two categories: task-agnostic and task-specific distillation \cite{sun-etal-2020-mobilebert}. Task-agnostic methods aim to perform distillation on the pre-training objective like masked language modeling (MLM) in order to obtain a smaller pre-trained model. However, many tasks of practical interest to the NLP community are not as complex as the MLM task solved by task-agnostic approaches. This results in complexity inversion --- i.e., in order to solve a specific relatively easy problem the models learn to solve a general much harder problem which entails language understanding. Task-specific methods on the other hand distill the knowledge needed to solve a specific task onto a student model thereby making the student very efficient at the task that it aims to solve. This allows for decoupled evolution of both the teacher and student models.

In task-specific distillation, the most important requirement is that the student model architectures are efficient in terms of the number of training samples, number of parameters, and number of FLOPS. To address this need we propose pQRNN (projection Quasi-RNN), an embedding-free neural encoder for NLP tasks. Unlike embedding-based model architectures used in NLP \cite{DBLP:journals/corr/WuSCLNMKCGMKSJL16, DBLP:journals/corr/VaswaniSPUJGKP17}, pQRNN learns the tokens relevant for solving a task directly from the text input similar to \citet{kaliamoorthi-etal-2019-prado}. Specifically, they overcome a significant bottleneck of multilingual pre-trained language models where embeddings take up anywhere between 47\% and 71\% of the total parameters due to large vocabularies~\cite{chung2020rethinking}. This results in many advantages such as not requiring pre-processing before training a model and having orders of magnitude fewer parameters. 

Our main contributions in this paper are as follows
\vspace{-4pt}
\begin{enumerate}[itemsep=0.2pt]
\item We propose pQRNN -- a tiny, efficient, and embedding-free neural encoder for NLP tasks.
\item We demonstrate the effectiveness of pQRNNs as a student architecture for task-specific distillation of large pre-trained language models.
\item We propose a simple and effective strategy for data augmentation which further improves quality and perform careful ablations on pQRNN parameters and distillation settings.
\item We show in experiments on two semantic-parsing datasets, that our pQRNN student is able to achieve $>$95\% of the teacher performance while being 350x smaller than the teacher.
\end{enumerate}

\section{Related Work}
\textbf{Distillation of pre-trained models} In task-specific distillation, the pre-trained model is first fine-tuned for the task and then distilled. \citet{DBLP:journals/corr/abs-1903-12136} distill BERT into a simple LSTM-based model for sentence classification. In \citet{sun-etal-2019-patient}, they also extract information from the intermediate layers of the teacher. \citet{chatterjee2019making} distills a BERT model fine-tuned on SQUAD into a smaller transformer model that's initialized from BERT. \citet{tsai2019small} distills mBERT into a single multilingual student model which is 6x smaller on sequence-labeling tasks. In task-agnostic distillation, the distillation is performed at the pre-training stage. Some examples of this strategy are: MobileBERT \cite{sun-etal-2020-mobilebert} which distills BERT into a slimmer version and MiniLM \cite{wang2020minilm} distills the self-attention component during pre-training. In TinyBERT~\cite{jiao2020tinybert} and \citet{turc2019wellread}, the authors perform a combination of the above two strategies where they perform distillation both at the pre-training and fine-tuning stages. \newline
\textbf{Model Compression} Apart from knowledge distillation, there has been an extensive body of work around compressing large language models. Some of the most prominent methods include: low-rank approximation~\cite{lan2020albert}, weight-sharing~\cite{lan2020albert, dehghani2018universal}, pruning~\cite{cui-etal-2019-fine, gordon2020compressing, hou2020dynabert}, and quantization~\cite{shen2019qbert, zafrir2019q8bert}.

\section{Proposed Approach}
\subsection{Overview}
We perform distillation from a pre-trained mBERT teacher fine-tuned for the semantic parsing task. We propose a projection-based architecture for the student. We hypothesise that since the student is task-specific, using projection would allow the model to learn the relevant tokens needed to replicate the decision surface learnt by the teacher. This allows us to significantly reduce the number of parameters that are context invariant, such as those in the embeddings, and increase the number of parameters that are useful to learn a contextual representation. We further use a multilingual teacher that helps improve the performance of low-resource languages through cross-lingual transfer learning. We propose input paraphrasing as a strategy for data augmentation which further improves the final quality. 
\subsection{Data augmentation strategy}
Large pre-trained language models generalize well beyond the supervised data on which they are fine-tuned. However we show that the use of supervised data alone is not effective at knowledge transfer during distillation. Most industry scale problems have access to a large corpus of unlabeled data that can augment the supervised data during distillation. However, this is not true for public datasets. We leverage query paraphrasing through bilingual pivoting~\cite{mallinson-etal-2017-paraphrasing} as a strategy to overcome this problem. This is achieved by using a in-house neural machine translation system~\cite{DBLP:journals/corr/WuSCLNMKCGMKSJL16, johnson-etal-2017-googles} to translate a query in the source language $S$ to a foreign language $T$ and then back-translating it into the original source language $S$, producing a paraphrase. We show that this introduces sufficient variation and novelty in the input text and helps improve knowledge transfer between the teacher and the student.
 
\subsection{Teacher Model Architecture: mBERT}
We used mBERT~\cite{DBLP:journals/corr/abs-1810-04805}, a transformer model that has been pretrained on the Wikipedia corpus of 104 languages using MLM and Next Sentence Prediction (NSP) tasks, as our teacher. The model architecture has 12 layers of transformer encoder \cite{DBLP:journals/corr/VaswaniSPUJGKP17} with a model dimension of 768 and 12 attention heads. It uses an embedding table with a wordpiece~\cite{schuster2012japanese} vocabulary of 110K resulting in approximately 162 million parameters. The mBERT teacher is then fine-tuned on the multilingual training data for the relevant downstream task. This results in a single multilingual task-specific teacher model for each dataset.
\subsection{Student Model Architectures: pQRNN}
\begin{figure}[t]
    \centering
    \includegraphics[trim={5cm 19cm 5cm 2.5cm},clip,width=8cm]{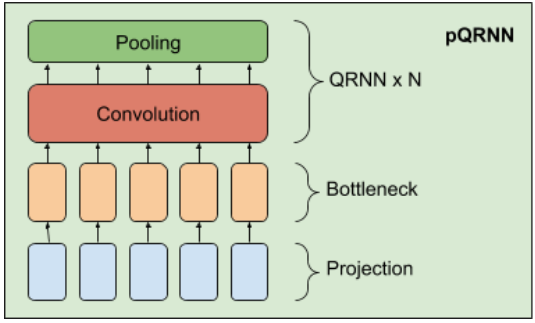}
    \caption{The pQRNN model architecture}
    \label{fig:pqrnn}
\end{figure}

The pQRNN model architecture, shown in Figure \ref{fig:pqrnn}, consists of three distinct layers which we describe below.
\subsubsection{Projection Layer} We use the projection operator proposed in \cite{kaliamoorthi-etal-2019-prado} for this work. The operator takes a sequence of tokens and turns it into a sequence of ternary vectors $\in \mathbb{P}^N$, where $\mathbb{P}$ is the set $\{-1, 0, 1\}$ and N is a hyper-parameter. This is achieved by generating a fingerprint for each token with $2N$ bits and using a surjective function to map every two bits to $\mathbb{P}$. The surjective function maps the binary values $\{00, 01, 10, 11\}$ to $\{-1, 0, 0, 1\}$ and has a few useful properties. 1) Since the value $0$ is twice as likely as the non-zero values and the non-zero values are equally likely, the representation has a symmetric distribution. 2) For any two non-identical tokens the resulting vectors are orthogonal if the fingerprints are truly random. This can be proved by performing a dot product of the vectors $\in \mathbb{P}^N$ and showing that the positive and negative values are equally likely, and hence the expected value of the dot product is 0. 3) The norm of the vectors are expected to be $\sqrt{N/2}$ since $N/2$ values are expected to be non zero.

\subsubsection{Bottleneck Layer} The result of the projection layer is a sequence of vectors $\bold{X} \in  \mathbb{P}^{T \times N}$, where $T$ is the number of tokens in the input sequence. Though this is a unique representation for the sequence of tokens, the network has no influence on the representation and the representation does not have semantic similarity. So we combine it with a fully connected layer to allow the network to learn a representation useful for our task.
\[ReLU(BatchNorm(\bold{X}\bold{W} + b))\]
where $\bold{W} \in \mathbb{R}^{N \times B}$ and $B$ is the bottleneck dimension.
\subsubsection{Contextual Encoder} The result of the bottleneck layer is invariant to the context. We learn a contextual representation by feeding the result of the bottleneck layer to a stack of sequence encoders. In this study we used a stack of bidirectional QRNN \cite{bradbury2016quasirecurrent} encoders and we refer to the resulting model as
{\bf pQRNN}. We make a couple of modifications to the QRNN encoder.
\begin{enumerate}
\item We use batch normalization \cite{DBLP:journals/corr/IoffeS15} before applying {\it sigmoid} and {\it tanh} to the Z/F/O gates. This is not straight-forward since the inputs are zero padded sequences during training. We circumvented this issue by excluding the padded region when calculating the batch statistics. 
\item We use quantization-aware training \cite{DBLP:journals/corr/abs-1712-05877} to construct the model.
\end{enumerate}
The hyper-parameters of the model are the state size $S$, the kernel width $k$ and the number of layers $L$. As described in \cite{bradbury2016quasirecurrent}, we use zoneout for regularization. However, we use a decaying form of zoneout which sets the zoneout probability in layer $l \in [1, ..., L]$ as $b^l$, where $b$ is a base zoneout probability. This is done based on the observation that regularization in higher layers makes it harder for the network to learn a good representation since it has fewer parameters to compensate for it, while this is not the case in lower layers.
\section{Experimental Setup}
\subsection{Model Configuration: pQRNN}
We configure the pQRNN encoder to predict both intent and arguments as done in  \cite{li2020mtop}. But we decompose the probabilities using chain rule, as follows
\[ P(i, a_{1:T} | q) = P(a_{1:T} | i, q) P(i | q)\]
where $i$, $a_{1:T}$, and $q$ are the intent, arguments and query respectively. The first term in the decomposition is trained with teacher forcing \cite{teacher-forcing} for the intent values. The arguments are derived from the pQRNN outputs as follows
\[P(a_{1:T} | i, q) = Softmax(Linear(O) + W \delta_i)\]
where  $O \in \mathbb{R}^{T\times 2S} $ are the outputs of the bidirectional pQRNN encoder with state size $S$, $W \in \mathbb{R}^{A \times I}$ is a trainable parameter, where $I$ and $A$ are the number of intents and arguments respectively, $\delta_i \in \mathbb{R}^{I}$ is a one hot distribution on the intent label during training and most likely intent prediction during inference.  The intent prediction from the query is derived using attention pooling as follows
\[P(i | q) = Softmax(Softmax(W_{pool}O^T)O)\]
where $W_{pool} \in \mathbb{R}^{2S}$ is a trainable parameter.
\paragraph{Training Setup:} Using this model configuration we construct an optimization objective using cross-entropy loss on the intent and arguments. The objective includes an L2 decay with scale $1e^{-5}$ on the trainable parameters of the network. The training loss is optimized using Adam optimizer~\cite{kingma2017adam} with a base learning rate of $1e^{-3}$ and an exponential learning rate decay with decay rate of $0.9$ and decay steps of $1000$. Training is performed for 60,000 steps with a batch size of $4096$ and synchronous gradient updates \cite{DBLP:journals/corr/ChenMBJ16}. We evaluate the model continuously on the development set during training and the best model is later evaluated on the test set to obtain the metrics reported in this paper.
\label{sec:hyperparams}
\paragraph{Model Hyper-parameters:} The pQRNN model configuration we use in the experiments have a projection feature dimension $N = 1024$, bottleneck dimension $B = 256$, a stack of bidirectional QRNN encoder with number of layers $L = 4$, state size $S = 128$ and convolution kernel width $k = 2$. We used an open source projection operator \footnote{https://github.com/tensorflow/models/tree/master/research/seq\_flow\_lite} that preserves information on suffixes and prefixes. For regularization we use zoneout \cite{DBLP:journals/corr/KruegerMKPBKGBL16} with a layer-wise decaying probability of $0.5$. We also used dropout on the projection features with a probability of $0.8$. 
\subsection{Datasets}
We run our experiments on two multilingual semantic-parsing datasets: multilingual ATIS and MTOP. Both datasets have one intent and some slots associated with a query. Given a query, both intent and slots have to be predicted jointly.

\begin{table}[ht]
\centering
\resizebox{0.48\textwidth}{!}{
\begin{tabular}{l|rrrrr}
\toprule
\multirow{2}{*}{Language} & \multicolumn{3}{c}{Examples} & \multirow{2}{*}{Intents} & \multirow{2}{*}{Slots} \\
                          & train     & dev    & test    &                          &                        \\ \midrule
English                   & 4488      & 490    & 893     & 18                       & 84                     \\
Spanish                   & 4488      & 490    & 893     & 18                       & 84                     \\
Portuguese                & 4488      & 490    & 893     & 18                       & 84                     \\
German                    & 4488      & 490    & 893     & 18                       & 84                     \\
French                    & 4488      & 490    & 893     & 18                       & 84                     \\
Chinese                   & 4488      & 490    & 893     & 18                       & 84                     \\
Japanese                  & 4488      & 490    & 893     & 18                       & 84                     \\
Hindi                     & 1440      & 160    & 893     & 17                       & 75                     \\
Turkish                   & 578       & 60     & 715     & 17                       & 71                     \\ \bottomrule
\end{tabular}}
\caption{Data statistics for the MultiATIS++ corpus.}
\label{tab:atis_stat}
\end{table}

\paragraph{Multilingual ATIS:} It is a multilingual version of the classic goal oriented dialogue dataset ATIS \cite{price1990evaluation}. ATIS contains queries related to single domain i.e. air travel. ATIS was expanded to multilingual ATIS by translating to 8 additional languages \cite{xu2020end}. Overall, it has 18 intents and 84 slots. Table \ref{tab:atis_stat} shows the summary of the statistics of this dataset.

\begin{table}[ht]
\centering
\resizebox{0.48\textwidth}{!}{
\begin{tabular}{l|rrrrr}
\toprule
\multirow{2}{*}{Language} & \multicolumn{3}{c}{Examples} & \multirow{2}{*}{Intents} & \multirow{2}{*}{Slots} \\
                          & train    & dev     & test    &                          &                        \\ \midrule
English                   & 13152    & 1878    & 3757    & 113                     & 75                     \\
German                    & 11610     & 1658    & 3317     & 113                      & 75                     \\
French                    & 10821     & 1546     & 3092     & 110                      & 74                     \\
Spanish                   & 11327     & 1618     & 3236     & 104                      & 71                     \\
Hindi                     & 10637     & 1520     & 3039     & 113                      & 73                     \\
Thai                      & 13152     & 1879     & 3758     & 110                      & 73      \\
\bottomrule
\end{tabular}}
\caption{Data statistics for the MTOP corpus.}
\label{tab:mtop_stat}
\end{table}

\begin{table*}[h]
\centering
\resizebox{0.75\textwidth}{!}{
\begin{tabular}{l|c|cccccc|c}
\toprule
                              &              & \multicolumn{6}{c|}{Exact Match Accuracy} &       \\
                              & \#Params     & en   & es   & fr   & de   & hi   & th   & avg  \\
                \midrule
                \multicolumn{9}{c}{Reference} \\
\midrule
XLU          &    70M (float)          & 78.2 & 70.8 & 68.9 & 65.1 & 62.6 & 68.0   & 68.9 \\
XLM-R        &    550M (float)          & 85.3 & 81.6 & 79.4 & 76.9 & 76.8 & 73.8 & 79.0 \\
\midrule 
\multicolumn{9}{c}{Baselines} \\
\midrule
Transformer & 2M (float) &  71.7    &  68.2    &    65.1  & 64.1     & 59.1     &  48.4    &  62.8    \\
pQRNN       & 2M (8bit) & 78.8 & 75.1 & 71.9 & 68.2 & 69.3 & 68.4 & 71.9 \\
\midrule 
\multicolumn{9}{c}{Distillation} \\
\midrule
mBERT* teacher & 170M (float) & 84.4 & 81.8 & 79.7 & 76.5 & 73.8 & 72.0   & 78.0 \\
pQRNN student & 2M (8bit) & 81.8 & 79.1 & 75.8 & 70.8 & 72.1 & 69.5 & 74.8 \\
\midrule
\multicolumn{2}{c|}{Student/Teacher (\%)} & 96.9 & 96.7 & 95.1 & 92.5 & 97.7 & 96.5 & 95.9 \\
\bottomrule
\end{tabular}}
\caption{References, Baselines, Teacher and Student for MTOP dataset. Reference numbers have been taken from \cite{li2020mtop}. We use exact match accuracy as metric to compare w/references. }
\label{tab:mtop}
\end{table*}

\paragraph{MTOP:} It is a multilingual task-oriented semantic parsing
dataset covering 6 languages \cite{li2020mtop}. It contains a mix of both simple as well as compositional nested queries across 11 domains, 117 intents and 78 slots. Similar to multilingual ATIS, this dataset was also created by translating from English to five other target languages. The dataset provides both flat and hierarchical representations but for the purpose of our work, we only use the former.  Table \ref{tab:mtop_stat} shows a summary of the MTOP dataset.
\subsection{Data Augmentation}
It has been shown \cite{das2020empirical} that generalization performance of deep learning models can be improved significantly by simple data augmentation strategies. It is particularly useful for our case since we are trying to replicate a function learnt by a teacher model that has been trained on much more data than the student.

Our method for data augmentation uses round-trip translation through a pivot language which is different from that of the original query. To get a new utterances/queries, we take the original query from the dataset and first translate it to the pivot language using our in-house translation system. The query in the pivot language is then translated back to the original language. The round-trip translated queries are now slightly different from the original query. This increases the effective data size. The round-trip translated queries sometimes also have some noise which has an added advantage as it would make the model, which uses this data, robust to noise.

Once we get the queries, we obtain the intent and argument labels for these new queries by doing a forward pass through the teacher model that was trained on original queries. For our experiments in this paper, we run 3 variations: one where the number of augmented queries is equal to the size of the original data and two other variations where the augmented data is 4x or 8x the size of the original data.

\begin{table*}[ht]
\centering
\resizebox{0.9\textwidth}{!}{
\begin{tabular}{l|c|cccccccccc}
\toprule
                       &              & \multicolumn{10}{c}{Intent Accuracy}                               \\
                       & \#Params     & en   & de   & es   & fr   & hi   & ja   & pt   & tr   & zh   & avg  \\ \midrule
                \multicolumn{12}{c}{Reference} \\
\midrule
LSTM               &              & 96.1 & 94.0 & 93.0 & 94.7 & 84.5 & 91.2 & 92.7 & 81.1 & 92.5 & 91.1 \\
mBERT       &    170M (float)          & 95.5 & 95.1 & 94.1 & 94.8 & 87.8 & 92.9 & 94.0 & 85.4 & 93.4 & 92.6 \\
\midrule 
\multicolumn{12}{c}{Baselines} \\
\midrule
Transformer &    2M (float)          &  96.8    &   93.2   &   92.1   &   93.1   &   79.6   &   90.7   &   92.1   &  78.3    &  88.1    &  89.3     \\
           pQRNN       & 2M (8bit) & 98.0 & 96.6 & 97.0 & 97.9 & 90.7 & 88.7 & 97.2 & 86.2 & 93.5 & 94.0 \\ \midrule 
\multicolumn{12}{c}{Distillation} \\
\midrule
mBERT* teacher & 170M (float) & 98.3 & 98.5 & 97.4 & 98.6 & 94.5 & 98.6 & 97.4 & 91.2 & 97.5 & 96.9 \\
pQRNN student  & 2M (8bit) &      & 97.3     &      &      & 91.0      &      &      &   88.5   &      &      \\ \midrule
\multicolumn{2}{c|}{Student/Teacher (\%)}                                                                                                                                                    & \multicolumn{1}{l}{} & \multicolumn{1}{l}{98.0} & \multicolumn{1}{l}{} & \multicolumn{1}{l}{} & \multicolumn{1}{l}{96.3} & \multicolumn{1}{l}{} & \multicolumn{1}{l}{} & \multicolumn{1}{l}{97.0} & \multicolumn{1}{l}{} & \multicolumn{1}{l}{} \\
\midrule
\end{tabular}
}
\caption{References, Baselines, Teacher and Student for multilingual ATIS dataset. Reference numbers have been taken from \cite{xu2020end}. We report both intent accuracy and argument f1 metrics to compare w/references. }
\label{tab:atis}
\end{table*}

\section{Results \& Ablations}
\subsection{MTOP Results}
\label{sec:mtop_results}
Table \ref{tab:mtop} presents our results on the MTOP dataset. The \textit{Reference} subsection presents results from prior work. XLU refers to a bidirectional LSTM-based intent slot model as proposed in \citet{Liu+2016, 10.5555/3060832.3061040} using pre-trained XLU embeddings. XLM-R~\cite{conneau-etal-2020-unsupervised} refers to a large multilingual pre-trained model with 550M parameters. From the \textit{Baselines} subsection, we can see that the pQRNN baseline trained just on the supervised data significantly outperforms a comparable transformer baseline by $9.1\%$ on average. Our pQRNN baseline also significantly beats the XLU biLSTM pre-trained model by 3\% on average despite being 140x smaller. In the \textit{Distillation} subsection, we present our mBERT teacher results which are slightly worse than the XLM-R reference system since mBERT has far fewer parameters compared to XLM-R. For building the student, we experimented with distillation on just the supervised data and with using 4x or 8x additional paraphrased data. The best result from these three configurations is presented as \textit{pQRNN student}. It can be observed that on average our distillation recipe improves the exact match accuracy by close to 3\% when compared to the pQRNN baseline. Furthermore, our student achieves 95.9\% of the teacher performance despite being 340x smaller.

\subsection{Multilingual ATIS Results}
Table \ref{tab:atis} showcases the results for Multilingual ATIS dataset. For simplicity, we report just the intent accuracy metric in Table~\ref{tab:atis} and show the argument F1 metric in Table~\ref{tab:atis_arg} in the Appendix. The \textit{Reference} subsection has LSTM-based results presented in \cite{xu2020endtoend} and our own mBERT baseline. As expected, our mBERT system is significantly better than the LSTM system presented in \cite{xu2020endtoend}. We see similar trends as seen in section~\ref{sec:mtop_results} on this dataset as well. Our pQRNN baseline significantly outperforms a comparable transformer baseline by avg. 4.3\% intent accuracy. Furthermore, the pQRNN baseline performs significantly better than the LSTM reference system by avg. 2.9\%. It should be noted that the pQRNN baseline outperforms the mBERT reference system by avg. 1.4\% on intent accuracy. This is significant since pQRNN has 85x fewer parameters. In the \textit{Distillation} subsection, we present the results of our multilingual mBERT teacher which serves as the quality upper-bound. We can see that the headroom between the teacher and the pQRNN baseline is low for the following languages: en, es, fr, and pt. Note that ja and zh are not space-separated languages, thereby requiring access to the original segmenter used in the multilingual ATIS dataset to segment our paraphrased data and distill them using the teacher. Since this is not publicly available, we skip distillation on these two languages. This leaves us with distillation results on de, hi and tr. From the Table, we can see that we obtain on average 97.1\% of the teacher performance while being dramatically smaller.


\begin{table*}[!htbp]
\centering
\begin{tabular}{l|ccccccccc}
\toprule
 & D & 1 & 2 & 3 & 4 & 5 & 6 & 7 \\
\midrule
Quantized           & Yes  & No   &                      &      &      &      &      &      \\
Batch Norm          & Yes  &      & No                   &      &      &      &      &      \\
Balanced            & Yes  &      &                      & No   &      &      &      &      \\
Zoneout Probability & 0.5  &      &                      &      & 0.0  &      &      &      \\
State Size          & 128  &      &                      &      &      & 96   &      &      \\
Bottleneck          & 256  &      &                      &      &      &      & 128  &      \\
Feature Size        & 1024 &      &                      &      &      &      &      & 512  \\
\midrule
Intent Accuracy     & 92.7 & 92.6 & 90.8                 & 91.7 & 89.2 & 92.2 & 91.3 & 92.1 \\
Argument F1         & 81.9 & 81.9 & 79.1                 & 80.5 & 77.9 & 80.4 & 80.6 & 81.0 \\
Exact Match         & 68.2 & 68.2 & 64.3                 & 67.1 & 61.1 & 66.4 & 66.8 & 67.9 \\
\bottomrule
\end{tabular}
\caption{The table shows the impact of various hyper-parameters on the pQRNN model performance evaluated on the de language in MTOP dataset. D is the default setting and in each of the ablations, we change exactly one hyper-parameter.}
\label{tab:ablation}
\end{table*}

\subsection{pQRNN Ablation}
In this section, we study the impact of various hyper-parameters on the pQRNN model performance. We use the German MTOP dataset for this study. We report intent accuracy, slot F1 and exact match accuracy metrics for the baseline configuration in Table \ref{tab:ablation}.

We first studied the impact quantization had on the results while keeping all other parameters unchanged. From the Table, we can observe that quantization acts as a regularizer and helps improve the performance albeit marginally. Furthermore, it can be observed that disabling batch normalization results in a very significant degradation of the model performance. Next, we studied the impact of the mapping function. Instead of using a balanced map that results in symmetric and orthogonal projections, we created projection that had positive value twice as likely as the non-positive values; this degrades the performance noticeably. Setting the zoneout probability to 0.0, has the most significant impact on the model performance. We believe that due to the small dataset size, regularization is essential to prevent over-fitting. Finally, we study the impact of the hyper-parameters such as $N, B, S$ defined in Section \ref{sec:hyperparams} on the model performance by halving them. It can be observed that of the three parameters halving $S$ degrades the model performance the most especially for slot tagging. This is followed by $B$ and then $N$.

\subsection{Effect of Data Augmentation}

\begin{table}[!hptb]
\centering
\small
\begin{tabular}{l|cccccc}
\toprule
                     & en & de & es & fr & hi & th   \\
                     \midrule
teacher & 84.4 & 76.5 & 81.8 & 79.7 & 73.8 & 72.0 \\
                     \midrule
1x data & 79.4 & 68.6 & 75.4 & 73.0 & 70.2 & 69.5 \\
4x data & 81.2 & 70.6 & 78.5 & 75.8 & 72.1 & 63.9 \\
8x data & 81.8 & 70.8 & 79.1 & 74.9 & 71.2 & 62.7 \\
\bottomrule
\end{tabular}
\caption{Effect of data augmentation on student's performance on MTOP. The metric used is exact match accuracy.}
\label{tab:augmentation}
\end{table}

\begin{figure*}[ht]
    \centering
    \includegraphics[clip,width=10cm]{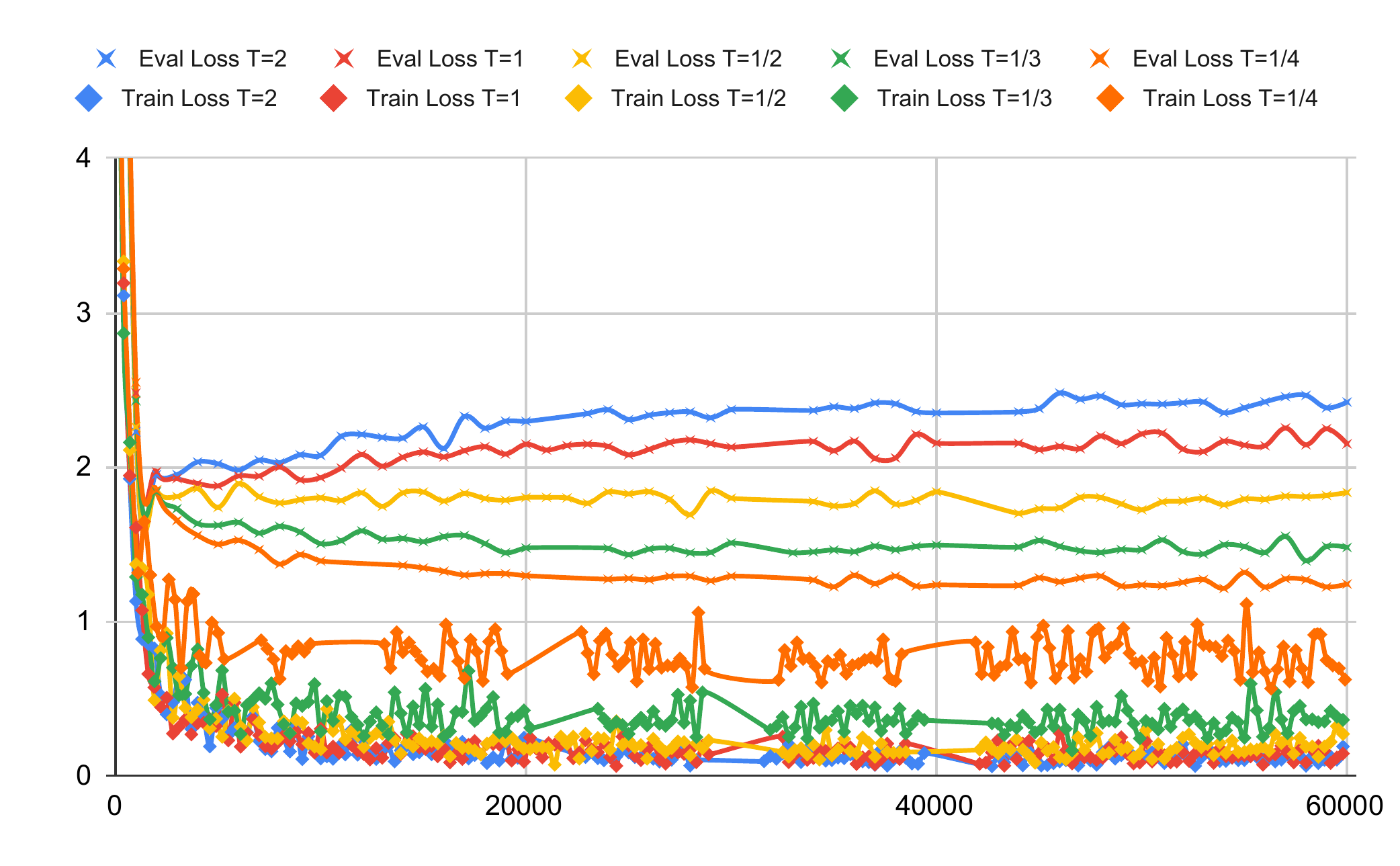}
    \caption{The effect of changing the scale of the teacher logits on the training and evaluation loss. The figure shows the training and eval loss in vertical axis against the training steps in horizontal axis.}
    \label{fig:temperature}
\end{figure*}

Table \ref{tab:augmentation} captures the effect of the data size on distillation. It shows the exact match accuracy for all languages on the MTOP dataset when we perform distillation on just the supervised training data as well as when we augment the training data with four times and eight times the data using the paraphrasing strategy. The paraphrasing strategy uses different pivot languages to generate the samples hence noise in the data is not uniform across all augmented samples. It can be observed that for all languages except Thai, when we use four times the data, the model performance improves. We believe the main reason for this is that the paraphrasing technique results in utterances that include different ways to express the same queries using novel words not part of the training set. The student pQRNN model learns to predict the classes with these novel words and hence generalizes better. For Hindi and French, as we increase the size further to eight times the original dataset we observe a slight drop in performance. We believe this is mostly due to noise introduced in the paraphrased data such as punctuation that affects the generalization of the BERT teacher. However for Thai, the distillation performs worse as we augment data. Further analysis revealed that the paraphrased data was segmented differently since we didn't have access to the segmenter used to create the original dataset. This introduces a mismatch between the supervised and the paraphrased data which results in poor distillation performance. These findings indicate that one could achieve even better performance if they have access to a large corpus of unlabeled data from the input domain and have full control over the segmenters used during supervised data creation.

\subsection{Effect of scaling the teacher logits}

\begin{figure}[h]
    \centering
    \includegraphics[trim={0.5cm 0cm 0cm 0cm}, clip,width=7cm]{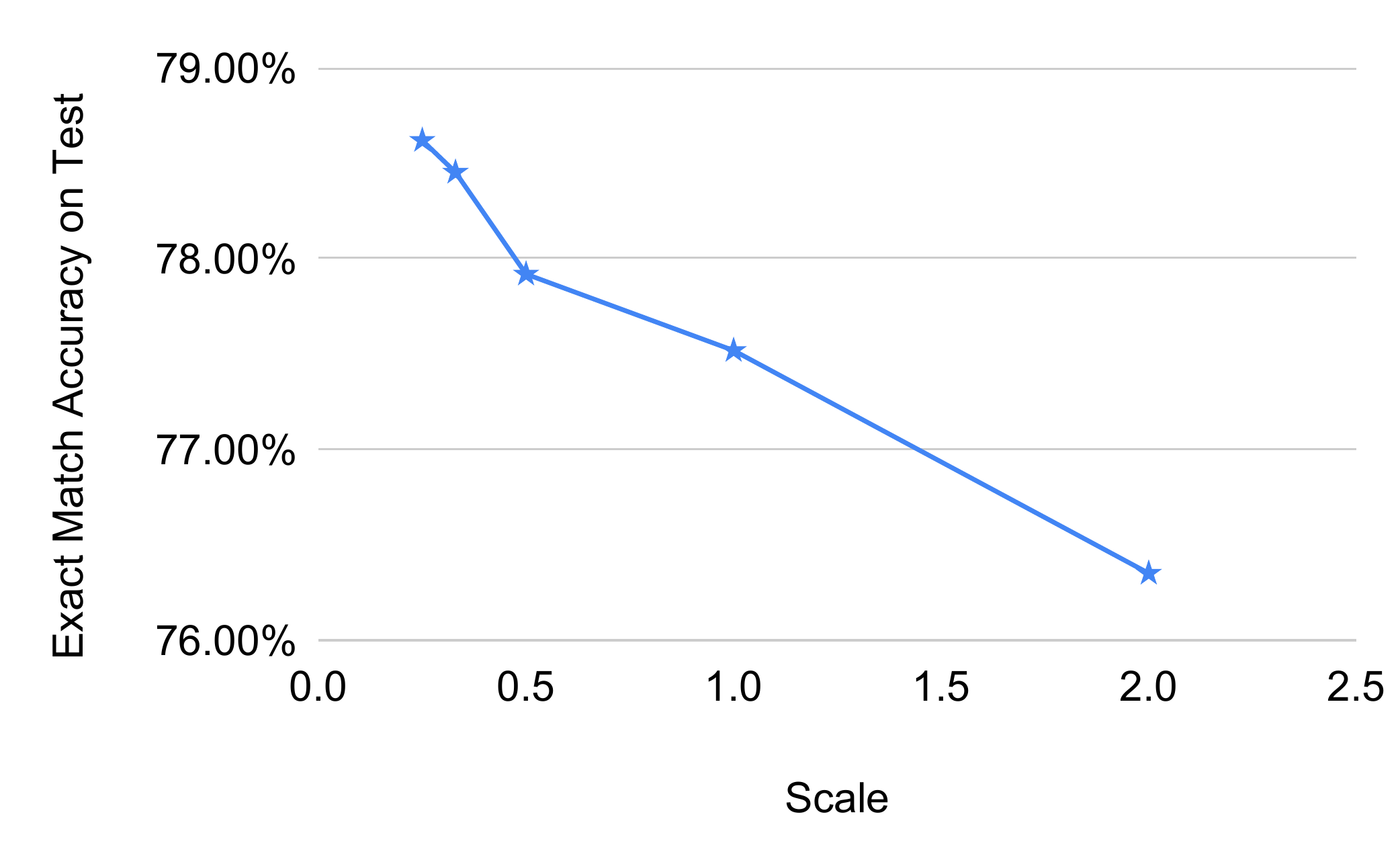}
    \caption{The effect of changing the scale of teacher logits on the exact matching accuracy on the test set.}
    \label{fig:scaleperformance}
\end{figure}

In this section, we study the effect of scaling the teacher logits on the MTOP Spanish dataset. This is not the same as distillation temperature discussed in \cite{hinton2015distilling}, since we change the temperature of just the teacher model. We plot the eval and train loss for training runs using different scales in Figure \ref{fig:temperature}. Scaling the logits with a multiplier greater than 1.0 results in a more peaky distribution whereas using a multiplier less than 1.0 result in a softer distribution for the intent and slot prediction probabilities from the teacher. It can be observed that as the scalar multiple is reduced the eval loss reduces and remains low as the training progresses. However, with higher temperatures (e.g., $T={1,2}$)  the eval loss first reduces and then increases indicating over-fitting. From these observations, we conclude that the exact match accuracy on the test set improves as the scale of the logits is reduced. This is further demonstrated in Figure \ref{fig:scaleperformance}.


\section{Conclusion}
We present pQRNN: a tiny, efficient, embedding-free neural encoder for NLP tasks. We show that pQRNNs outperform LSTM models with pre-trained embeddings despite being 140x smaller. They are also parameter efficient which is proven by their gain over a comparable transformer baseline. We then show that pQRNNs are ideal student architectures for distilling large pre-trained language models (i.e., mBERT). On MTOP, a multilingual task-oriented semantic parsing dataset, pQRNN students reach 95.9\% of the mBERT teacher performance. Similarly, on mATIS, a popular semantic parsing task, our pQRNN students achieve 97.1\% of the teacher performance. In both cases, the student pQRNN is a staggering 350x smaller than the teacher. Finally, we carefully ablate the effect of pQRNN parameters, the amount of pivot-based paraphrasing data, and the effect of teacher logit scaling. Our results prove that it's possible to leverage large pre-trained language models into dramatically smaller pQRNN students without any significant loss in quality. Our approach has been shown to work reliably at Google-scale for latency-critical applications.

\section*{Acknowledgments}

We would like to thank our colleagues Catherine Wah, Edgar Gonzàlez i Pellicer, Evgeny Livshits, James Kuczmarski, Jason Riesa and Milan Lee for helpful discussions related to this work. We would also like to thank Amarnag Subramanya, Andrew Tomkins, Macduff Hughes and Rushin Shah for their leadership and support.

\bibliography{anthology,acl2020}
\bibliographystyle{acl_natbib}

\appendix
\section{Appendix}
For the multilingual ATIS dataset, we present Argument F1 metrics in table~\ref{tab:atis_arg} below.

\begin{table*}[h]
\centering
\resizebox{0.9\textwidth}{!}{
\begin{tabular}{l|c|cccccccccc}
                      &              & \multicolumn{10}{c}{Argument F1}                               \\
                      & \#Params     & en   & de   & es   & fr   & hi   & ja   & pt   & tr   & zh   & avg  \\ \midrule
                \multicolumn{12}{c}{Reference} \\
\midrule
LSTM              &              & 94.7 & 91.4 & 75.9 & 85.9 & 74.9 & 88.8 & 88.4 & 64.4 & 90.8 & 83.9  \\
mBERT      &       170M (float)       & 95.6 & 95.0 & 86.6 & 89.1 & 82.4 & 91.4 & 91.4 & 75.2 & 93.5 & 88.9 \\
\midrule 
\multicolumn{12}{c}{Baselines} \\
\midrule
Transformer &   2M (float)          &    93.8  &   91.5   &  73.9    &   82.8   &   71.5   &  88.1    &  85.9    &   67.7   &   87.9   &  82.6    \\
          pQRNN       & 2M (8bit) & 95.1 & 93.9 & 87.8 & 91.3 & 79.9 & 90.2 & 88.8 & 75.7 & 92.2 & 88.3 \\ \midrule 
\multicolumn{12}{c}{Distillation} \\
\midrule
mBERT* teacher    & 170M (float) & 95.7 & 95.0 & 83.5 & 94.2 & 87.5 & 92.9 & 91.4 & 87.3 & 92.7 &  91.3 \\
pQRNN student      & 2M (8bit) &      & 94.8     &      &      & 84.5      &      &      &   77.7   &      &      \\ \midrule
\multicolumn{2}{c|}{Student/Teacher (\%)}                                                                                                                                                    & \multicolumn{1}{l}{} & \multicolumn{1}{l}{99.7} & \multicolumn{1}{l}{} & \multicolumn{1}{l}{} & \multicolumn{1}{l}{96.6} & \multicolumn{1}{l}{} & \multicolumn{1}{l}{} & \multicolumn{1}{l}{89.0} & \multicolumn{1}{l}{} & \multicolumn{1}{l}{} \\
\bottomrule

\end{tabular}
\caption{References, Baselines, Teacher and Student argument F1 metrics for multilingual ATIS dataset. Reference numbers have been taken from \cite{xu2020end}. Intent Accuracy metric is reported in \ref{tab:atis}}
\label{tab:atis_arg}
}
\end{table*}

\end{document}